%% file: main.tex
\documentclass[10pt,twocolumn,letterpaper]{article}

\usepackage{cvpr}              

\input{preamble}

\definecolor{cvprblue}{rgb}{0.21,0.49,0.74}
\usepackage[pagebackref,breaklinks,colorlinks,citecolor=cvprblue]{hyperref}

\def\paperID{31} 
\def\confName{CVPR}
\def\confYear{2024}

\title{Mitigating Bias Using Model-Agnostic Data Attribution}

\author{Sander De Coninck, Sam Leroux, Pieter Simoens \\
 IDLab, Department of Information Technology at Ghent University - imec\\
Technologiepark 126, B-9052 Ghent, Belgium \\
{\tt\small \{sander.deconinck, sam.leroux, pieter.simoens\}@ugent.be}
}

\begin{document}
\maketitle
\input{sec/0_abstract}    
\input{sec/1_intro}
\input{sec/2_rel_work}
\input{sec/3_region_classifier}
\input{sec/4_attribution}

\input{sec/5_conclusion}

\input{sec/ack}
{
    \small
    \bibliographystyle{ieeenat_fullname}
    \bibliography{main}
}


\end{document}

%% file: preamble.tex
\usepackage[dvipsnames]{xcolor}

\usepackage{subcaption}
\usepackage{algorithm}
\usepackage{tabularx}
\usepackage{algorithmic}
\usepackage{multirow}
\usepackage{comment}
\usepackage[accsupp]{axessibility}

\usepackage{array}
\newcolumntype{$}{>{\global\let\currentrowstyle\relax}}
\newcolumntype{^}{>{\currentrowstyle}}

%% file: sec/0_abstract.tex
\begin{abstract}
Mitigating bias in machine learning models is a critical endeavor for ensuring fairness and equity. In this paper, we propose a novel approach to address bias by leveraging pixel image attributions to identify and regularize regions of images containing significant information about bias attributes. Our method utilizes a model-agnostic approach to extract pixel attributions by employing a convolutional neural network (CNN) classifier trained on small image patches. By training the classifier to predict a property of the entire image using only a single patch, we achieve region-based attributions that provide insights into the distribution of important information across the image. We propose utilizing these attributions to introduce targeted noise into datasets with confounding attributes that bias the data, thereby constraining neural networks from learning these biases and emphasizing the primary attributes. Our approach demonstrates its efficacy in enabling the training of unbiased classifiers on heavily biased datasets.

\end{abstract}

%% file: sec/1_intro.tex
\section{Introduction}
\label{sec:intro}

\begin{figure}[ht]
    \centering
    \includegraphics[width=\columnwidth]{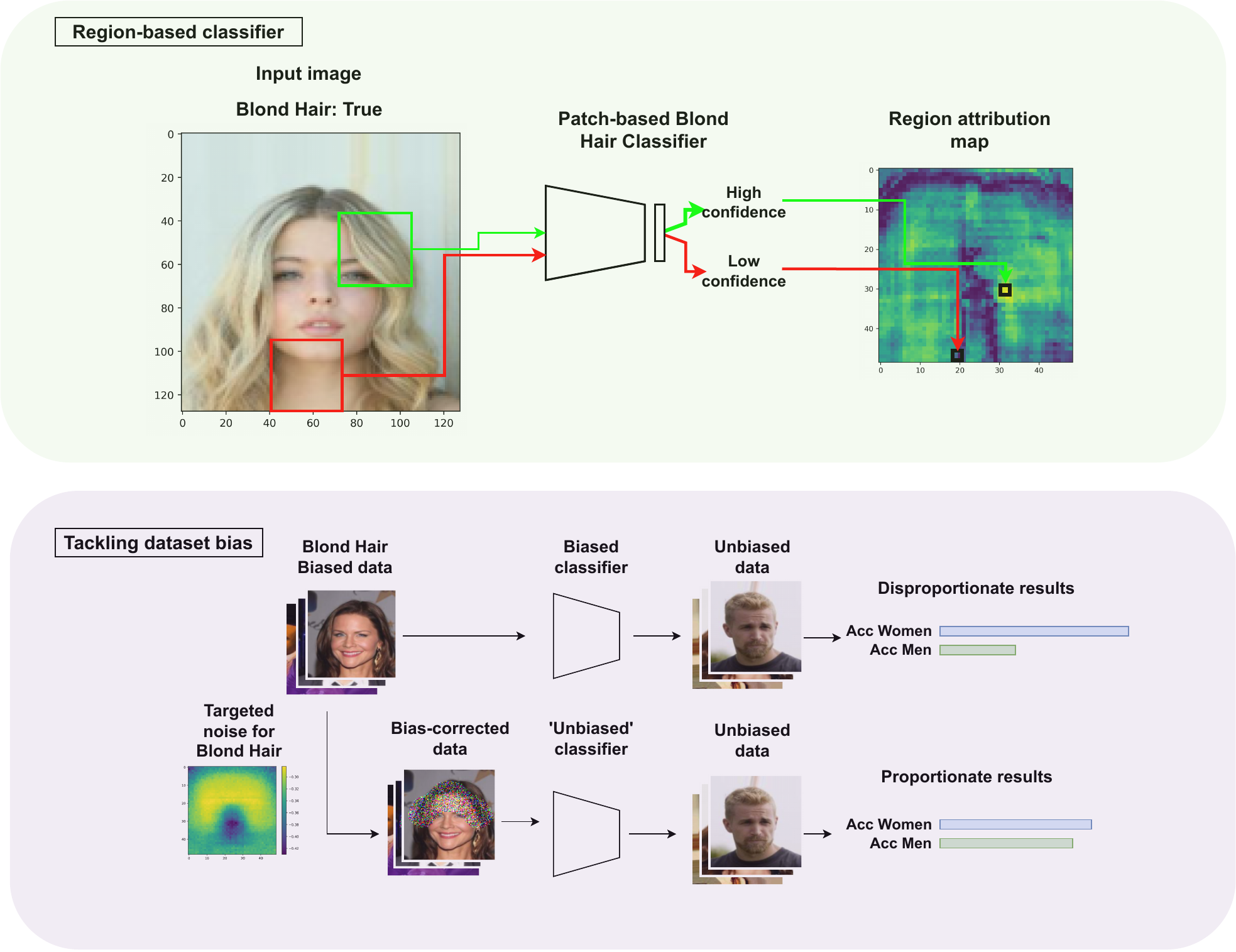}
    \caption{Our strategy to address biased learning. We utilize a region-based classifier to classify individual image patches based on attributes that introduce bias to the data. By leveraging the confidence of the trained model, we identify the regions within an image where attribute-related information is most concentrated. Subsequently, we introduce targeted noise to these areas in the training data to prevent models from overfitting to the confounding attributes.}
    \label{fig:region_classifier}
\end{figure}

In the domain of computer vision, various methodologies are employed to mitigate bias and uphold fairness in machine learning models operating on sensitive attributes. Data bias can manifest in various ways, such as when the data representation disproportionately favors a specific group of subjects \cite{olteanu2019social}. Alternatively, bias may arise from substantial differences between the training and production environments \cite{parraga2023fairness}. Moreover, biases within the data can be either explicit and acknowledged beforehand or concealed. We specifically focus on situations where bias is pre-existing and originates from disparities between the training and evaluation datasets due to a confounding variable heavily influencing the former. For instance, in the CelebA~\cite{liu2015faceattributes} dataset, a significant majority of women are depicted with blond hair, whereas only a minority of men exhibit this trait. Consequently, a model trained on such data may exhibit unequal performance across men and women in a production setting where this distribution does not hold true as it has learned to associate having blond hair with being a woman.

One prevalent technique to mitigate the learning of biases is dataset re-sampling, where instances from minority or majority classes are either over-sampled or under-sampled to achieve a more equitable distribution. However, in scenarios such as facial data, where numerous attributes are present, achieving a perfectly balanced dataset may be unattainable as this would require an equal number of samples for each subcategory. In the case of CelebA, which contains $40$ binary attributes, this would mean you require an equal amount of samples for each of the $2^{40}$ possible combinations. Another approach is to utilize dataset augmentation~\cite{parraga2023fairness}, which involves altering the training data to rectify imbalances in underrepresented classes or counter biases inherent in the training dataset, for example by generating synthetic data ~\cite{ramaswamy2021fair,yucer2020exploring}.  Our solution falls under this category as well. We propose a solution where we identify critical pixel areas for classifying attributes that introduce a bias in the data (e.g., blond hair in face data). By strategically introducing targeted noise to these regions in the training data, we aim to prevent the model from overfitting to these confounding attributes. A conceptualisation of our technique can be seen in Figure~\ref{fig:region_classifier}.

Attributing pixels that are important for classification is typically referred to as saliency mapping. Saliency mapping techniques were originally developed to elucidate machine learning models rather than datasets. Consequently, they attempt to identify pixel regions crucial for a specific model's classification. However, for mitigating biases being learned, it's more pertinent to determine which regions \textbf{potentially} contribute to classification. In other words, it's essential to discern which image regions harbour valuable information for a particular class (e.g., identifying facial features aiding in ethnicity classification). This will not always align with specific model attribution maps, as not all networks learn in the same way, and some may lay more importance on a certain pixel region than others.

To tackle this issue, we introduce data attribution through classification. We employ a small-scale classifier that predicts a sensitive label for the entire image from a single image patch. By identifying regions where the classifier confidently predicts the label, assuming the model is well calibrated, we establish a measure of information content regarding the sensitive label within that patch. As a result, this methodology allows us to attribute specific image regions to particular classes.

The resulting attributions could be used for a number of purposes, but we see the most promise in fields where it is undesirable that certain attributes are learned. The most notable of these is learning using a biased dataset. We propose to use our attributions as a way to regularize the training process, if some biases in the data are known a priori (e.g. due to a high correlation between labels), then we can compel the model learning to classify desired attributes to divert focus away from those regions important for classifying the confounder. 

We demonstrate the effectiveness of our attribution technique on two prominent face datasets, namely FairFace~\cite{karkkainen2021fairface} and CelebA~\cite{liu2015faceattributes}. These datasets are selected due to their comprehensive annotations and consistent alignment of posture, facilitating the validation of our attributions' utility through averaged attributions across multiple images. Subsequently, we explore the application of our attributions in training unbiased perceived gender classifiers on biased data, leveraging the CelebA dataset. By creating biased subsets where specific attributes are prominent for women but absent for men, we evaluate subgroup accuracy on a balanced dataset. Our experiments reveal that incorporating noise based on our attributions to the training data proves beneficial in mitigating the adverse effects of dataset bias on classifier performance.

We make the following contributions in this paper:
\begin{itemize}
    \item We developed a general model-independent data attribution technique that can foster data understanding
    \item We showcase the use of this attribution technique for data augmentation to train neural networks under known biases and experiment with different noise addition types and settings
\end{itemize}

The remainder of this paper is organized as follows. We position related works in Section \ref{sec:rel_work}. Following this, we introduce our region classifier and show how we obtain attributions in Section \ref{sec:region_classifier}. Lastly, we experiment with using the attributions for developing robust classifiers in Section \ref{sec:fair_training}. We conclude our work in Section \ref{sec:conclusion}. 

%% file: sec/2_rel_work.tex
\section{Related works}
\label{sec:rel_work}

\subsection{Dataset bias and fairness}

Correlation is a widely studied problem in fair research, mainly concerning spurious correlations in the dataset through societal biases or dataset construction. For example, for the CelebA dataset, it is known that the attractive attribute is biased towards women and that a majority of the women have blonde hair. Rajabi et al.~\cite{rajabi2023through} demonstrate that mitigating bias between gender and attractiveness for the CelebA dataset influences classification of several other attributes related to attractiveness as well. Similarly, Denton et al.~\cite{denton2019image} found that manipulating non-hair attributes such as heavy makeup resulted in a change in hair style/and or color.

Addressing these spurious correlations and biases represents an active area of investigation within the research community. Some studies concentrate on improving datasets to ensure a more equitable distribution of data samples, either through resampling techniques or augmenting the dataset with new data generated, for example, using generative models~\cite{ramaswamy2021fair,yucer2020exploring}. Others employ techniques during model training to guide the learning process, facilitating the acquisition or elimination of specific biases~\cite{zhang2018mitigating}. For instance, Kim et al. utilize a regularization loss based on mutual information to prevent the model from learning biased attributes~\cite{kim2019learning}. Additionally, certain approaches leverage causal learning methodologies to achieve unbiased recognition~\cite{wang2021causal}.

\subsection{Attribution beyond explainability}
Pixel attribution techniques are utilized to identify the areas or pixels within an image that significantly contribute to a model's prediction. These techniques typically fall into two categories: occlusion or perturbation-based, and gradient-based methods. The former, being model-agnostic, assesses the impact on prediction when certain regions are omitted, while the latter computes gradients concerning the input image~\cite{molnar2022}. Noteworthy gradient-based techniques include Grad-CAM~\cite{selvaraju2017grad}, Integrated Gradients~\cite{sundararajan2017axiomatic}, and XRAI~\cite{kapishnikov2019xrai}. For occlusion-based approaches, LIME~\cite{ribeiro2016should} and SHAP~\cite{lundberg2017unified} are widely recognized.

These methods primarily serve to enhance the interpretability of deep neural networks. However, they also see use beyond, in improving the generalization of deep learning models~\cite{Boyd_2022_WACV}, such as those used in facial expression recognition~\cite{khan2019saliency,mavani2017facial} and person detection tasks~\cite{aguilar2017cascade}, including scenarios involving thermal imagery~\cite{Ghose_2019_CVPR_Workshops}. Particularly relevant to our work, Huang et al.~\cite{huang2023gradient} employ saliency techniques to address biases in facial recognition models. Their approach employs gradient attention maps, ensuring consistent attention patterns across diverse racial backgrounds. However, our work differs in that we utilize saliency to mitigate confounding variables being learned, whereas they focus on ensuring uniform learning across different subgroups regardless of what is learned.

%% file: sec/3_region_classifier.tex
\section{Region classifier}
\label{sec:region_classifier}

In our efforts to alleviate learned biases, our focus is directed towards identifying the regions containing vital information for classification. While saliency techniques are commonly used for this purpose, they often focus solely on pinpointing important pixels for a specific model, potentially overlooking broader regions of interest. To address this, we deploy a region classifier capable of analyzing image patches and attributing characteristics of the entire image, such as determining a person's age. By assessing the confidence levels of a well-trained classifier, we gain valuable insights into the specific areas where significant information is concentrated. 

\subsection{Training setup}
\label{sec:training_setup}
Our region classifier employs a ResNet18 model architecture \cite{he2016deep}, which operates on patches of size $k \times k$. Additionally, the network takes in a region indicator, which indicates the region from which the patch was taken in the original image. This region indicator is translated into an embedding akin to the vision transformer approach \cite{dosovitskiy2020vit}. This embedding is then concatenated to the input image before execution. The embeddings and the classification network are trained simultaneously. During the training process, a single patch per image is randomly selected for training. Patches are randomly selected from $p$ possible positions, where patch $i$ corresponds to having $( (i\mod \sqrt{p}) * s, \lfloor i/\sqrt{p} \rfloor * s)$ as the top-left coordinates of the patch. Here, the stride $s$ is calculated as $I - k / \sqrt{p}$, with $I$ being the image size.

Our experiments utilize the FairFace \cite{karkkainen2021fairface} and CelebA~\cite{liu2015faceattributes} dataset, providing centred and aligned face images, which we resized to $128 \times 128$ pixels. We employ the AdamW optimizer \cite{loshchilov2017decoupled} with an initial learning rate of $10^{-3}$, reduced by a factor of $10$ every $40$ epochs during training, for a total of $90$ epochs. Patch dimensions $k$ are set to $32$, and the number of possible patch positions $p$ is $2041$. A batch size of $256$ is utilized, and the cross-entropy loss function with label smoothing of $0.1$ is employed.

\subsection{Confidence estimation}
To effectively utilize our region classifier for determining whether a patch contains relevant information regarding the subject, it is imperative that our classifier can provide an indication of its confidence level. There are three prevalent methods for estimating confidence when working with classification networks. Each relies on softmax values derived from the output logits $l$, which are computed as follows:

\begin{equation}
    S(l_{i}) = \frac{e^{l_i}}{\sum_{j}e^{l_j}}
\end{equation}

The three confidence indicators used are the top softmax score, the margin, and the negative entropy score, which are calculated as follows:

\begin{itemize}
    \item Top softmax score: this indicator is determined by selecting the highest softmax value among all classes.
    \begin{equation}
        \max_i s_i
    \end{equation}
    \item Margin: we determine it as the difference between the highest and second-highest softmax values.
    \begin{equation}
        \max_i s_i - \max_{j \neq \text{argmax}(s)} s_j
    \end{equation}
    \item Negative entropy score: it is computed as the negative sum of the softmax probabilities multiplied by their natural logarithms.
    \begin{equation}
         - \sum_i s_i \log s_i
    \end{equation}
\end{itemize}

In our experiments, we primarily rely on the negative entropy score as it considers the entire output distribution. While less straightforward to interpret than the top softmax score or the margin, we strive to enhance interpretability by normalizing outputs to a $[0, 1]$ range wherever feasible.

\subsection{Calibration}

When using our classifier for attribution, it's important that its confidence outputs are grounded in how accurately it can predict and, thus, how much information there is. A calibrated model is a model whose estimated confidence is close to its accuracy. The most common metrics of calibration are the Expected Calibration Error (ECE)~\cite{naeini2015obtaining}, and reliability diagrams~\cite{degroot1983comparison,niculescu2005predicting}.

Expected Calibration Error (ECE) quantifies the discrepancy between predicted and actual confidence levels. It is calculated as follows:

\begin{equation}
ECE = \sum^{N}_{i} b_i || (p_i - c_i) ||
\end{equation}

Here, $N$ represents the number of bins used for calibration, $b_i$ is the proportion of samples falling into bin $i$, $p_i$ denotes the average predicted confidence, and $c_i$ represents the average true accuracy. ECE measures the average difference between predicted confidence and actual accuracy across different confidence levels. It provides a holistic assessment of the calibration performance of a classifier.

Reliability diagrams, also known as calibration diagrams, graphically illustrate the alignment between predicted confidence levels and actual accuracy. They plot predicted confidence values against observed accuracy within equally spaced bins. Ideally, a well-calibrated classifier shows a diagonal line, indicating accurate confidence estimates. Deviations from this line highlight areas of overconfidence or underconfidence. Reliability diagrams offer a visual tool for evaluating a classifier's calibration performance.

\begin{figure*}[tb]
    \centering
    \begin{subfigure}{0.3\linewidth}
          \includegraphics[width=\linewidth]{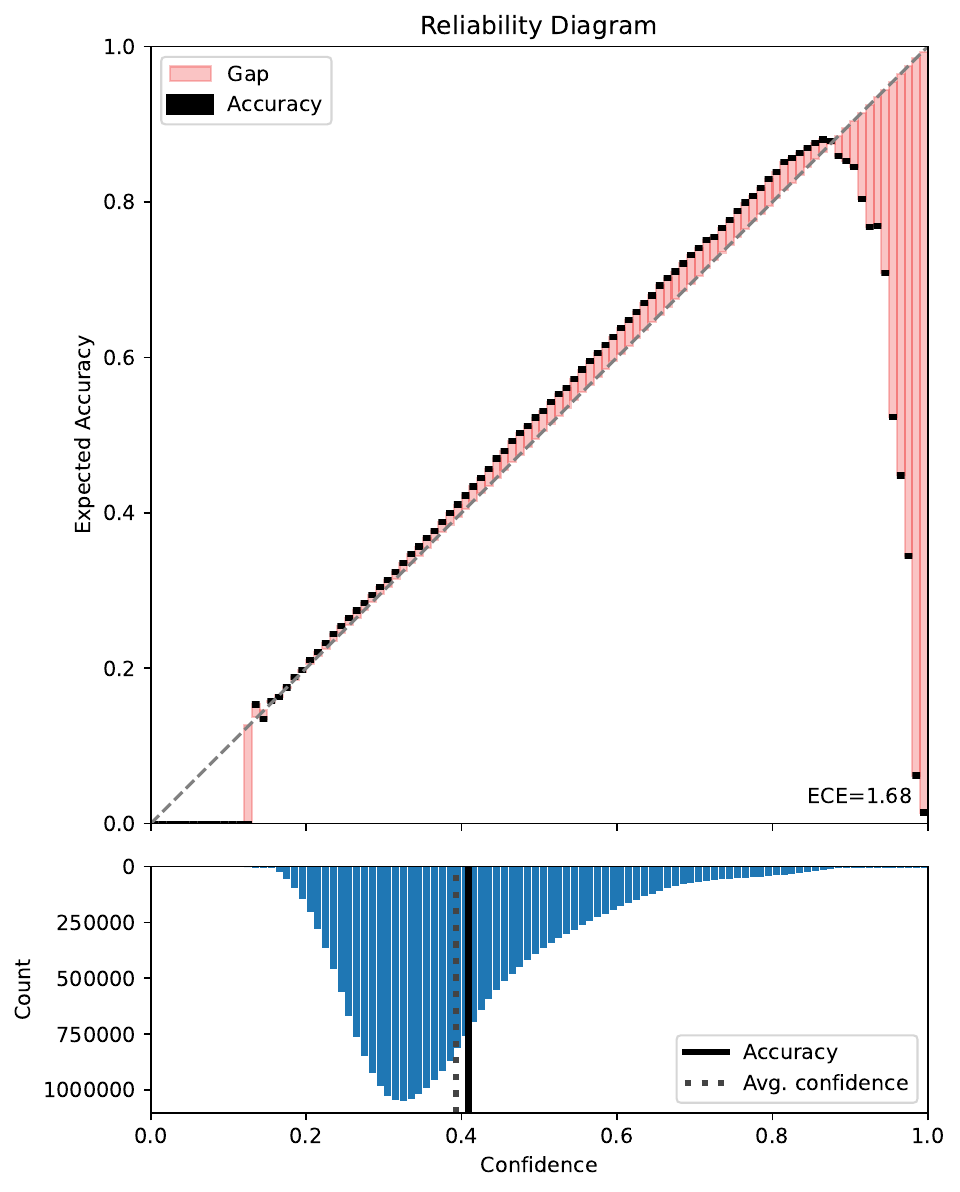}
    \caption{Age}
    \label{fig:rel_age}
    \end{subfigure}
    \begin{subfigure}{0.3\linewidth}
    \includegraphics[width=\linewidth]{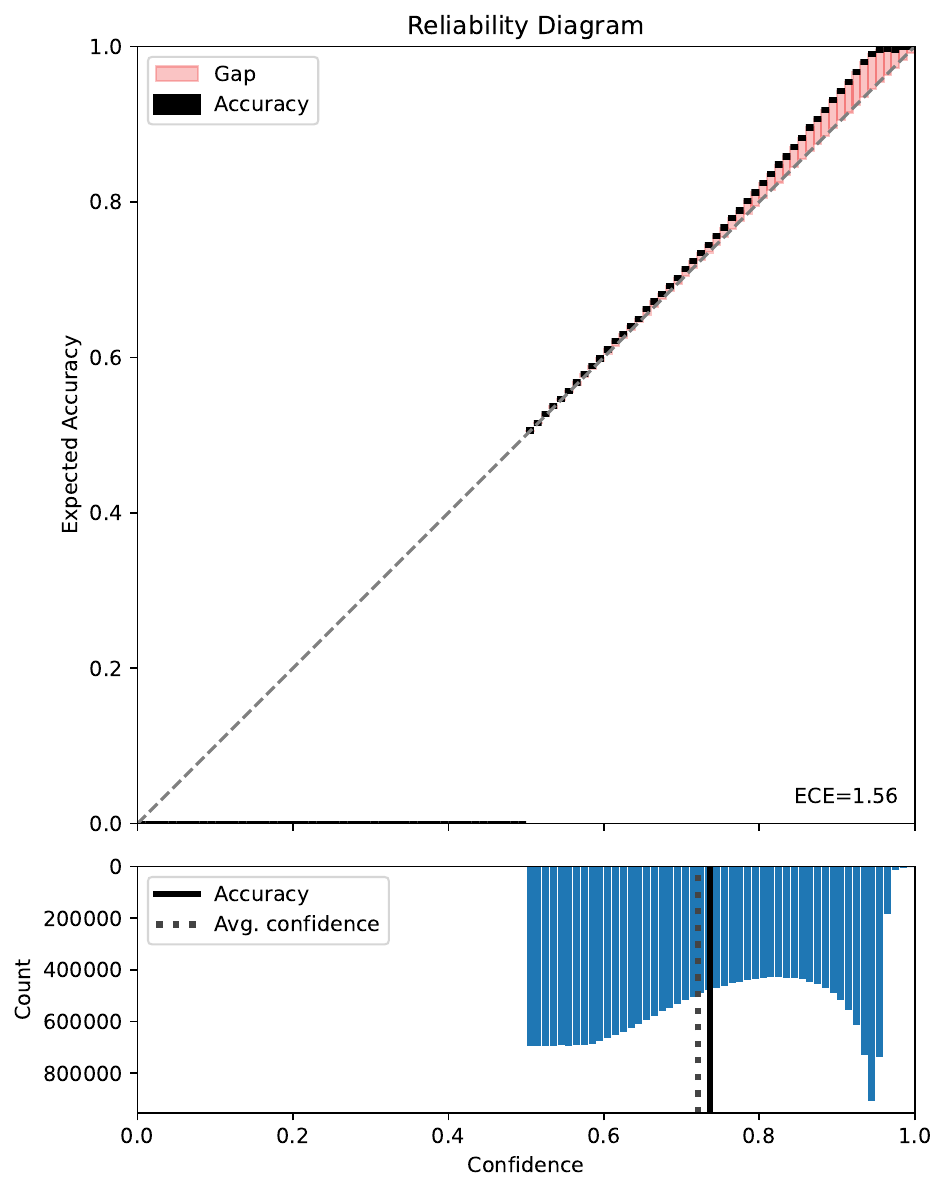}
    \caption{Gender}
    \label{fig:rel_gender}
    \end{subfigure}
    \begin{subfigure}{0.3\linewidth}
          \includegraphics[width=\linewidth]{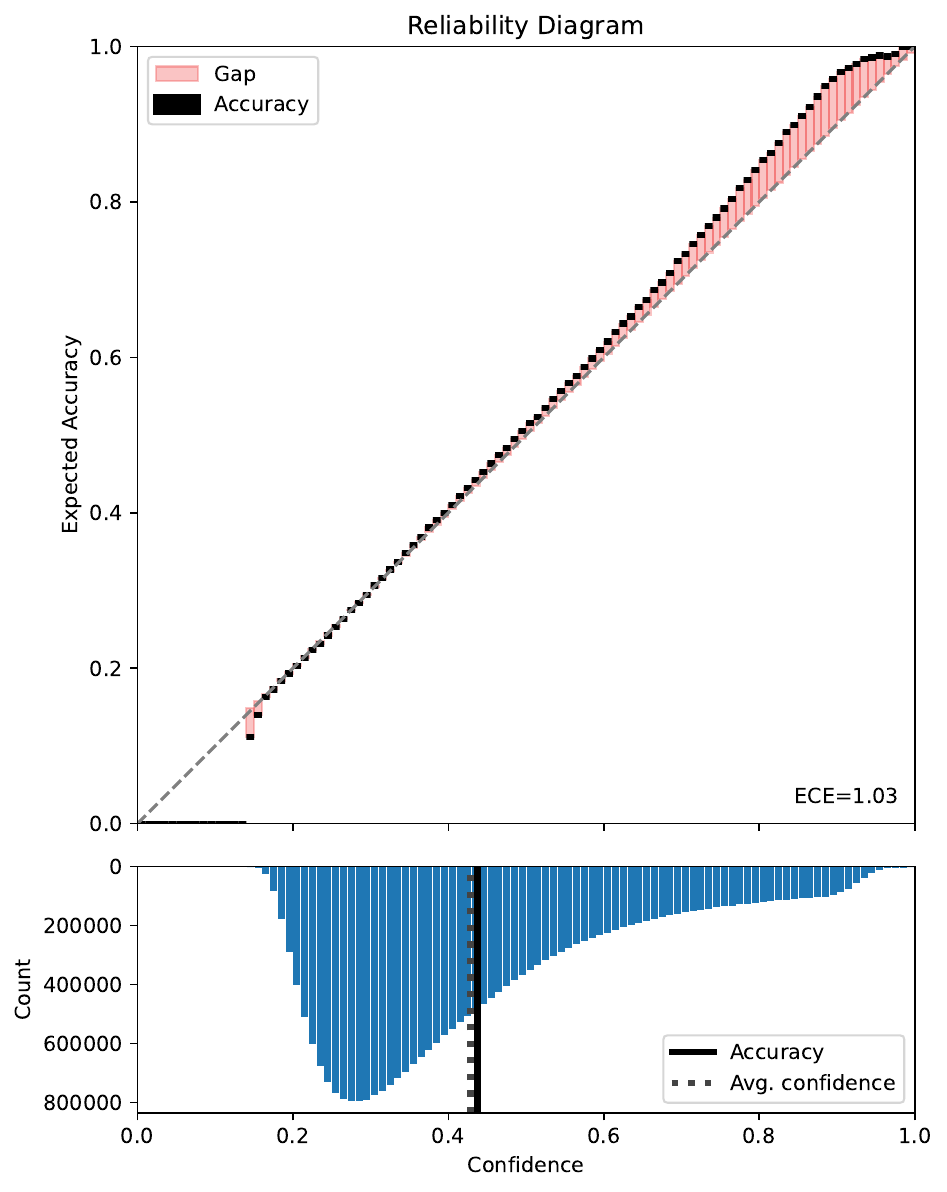}
    \caption{Race}
    \label{fig:rel_race}
    \end{subfigure}
    \caption{Reliability diagrams illustrating the performance of region classifiers trained on FairFace attributes. Our models exhibit strong calibration without any adjustments, as evidenced by the close alignment between average confidence and accuracy, alongside low Expected Calibration Error (ECE).}
    \label{fig:reliabity_diagrams}
\end{figure*}

Calibration diagrams, along with the corresponding Expected Calibration Error (ECE) scores, for networks trained to classify the three FairFace attributes are depicted in Figure \ref{fig:reliabity_diagrams}. These diagrams were generated using all possible patches from the validation set using $100$ bins. We observe that our networks are well-calibrated without having made any specific modifications. This is likely attributed to the high variability in data quality. Many regions may lack sufficient information for accurate classification but are still included, thereby mitigating overfitting and ensuring consistent calibration. Notably, for the age attribute, calibration appears to decrease significantly at high confidences, although such instances are infrequent, as evidenced by the lower plots. 

\subsection{Region-based attributions}

\begin{figure}[h]
\centering
\includegraphics[width=0.8\columnwidth]{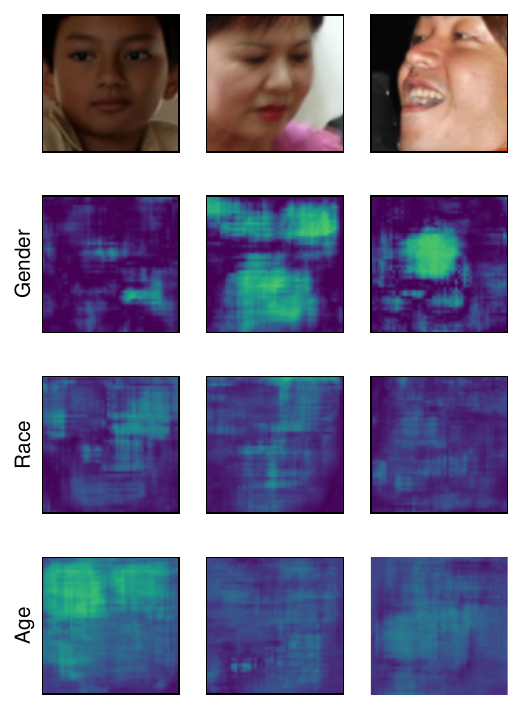}
\caption{Examples of region-based attributions.}
\label{fig:ex_reg}
\end{figure}

For the final step, we can use our patch-based classifiers, which we have shown to be well-calibrated, to calculate attribution maps. To attribute an image with respect to a specific attribute, we employ our region classifier to assess confidence across numerous regions. This is achieved by iteratively collecting regions in a sliding window manner, as can also be seen in Figure \ref{fig:region_classifier}. Subsequently, the obtained regions are batched and classified using the region classifier. Finally, a confidence score, such as negative entropy, is computed for each patch. The resulting attribution can be visualized as a $\sqrt{p} \times \sqrt{p} $ map, with each pixel $(x,y)$ representing a region whose top left coordinate in the original image is $(s*x,s*y)$. Illustrative examples of our region attributions for the FairFace dataset are showcased in Figure \ref{fig:ex_reg}. Note that we solely rely on confidence scores to calculate attributions, allowing a trained model to be applied to novel data without the need for ground-truth labels. One limitation of this attribution technique is its computational cost, as it requires classifying $p$ image patches per image.

\begin{figure}[]
    \centering
    \begin{subfigure}{0.45\linewidth}
          \includegraphics[width=\linewidth]{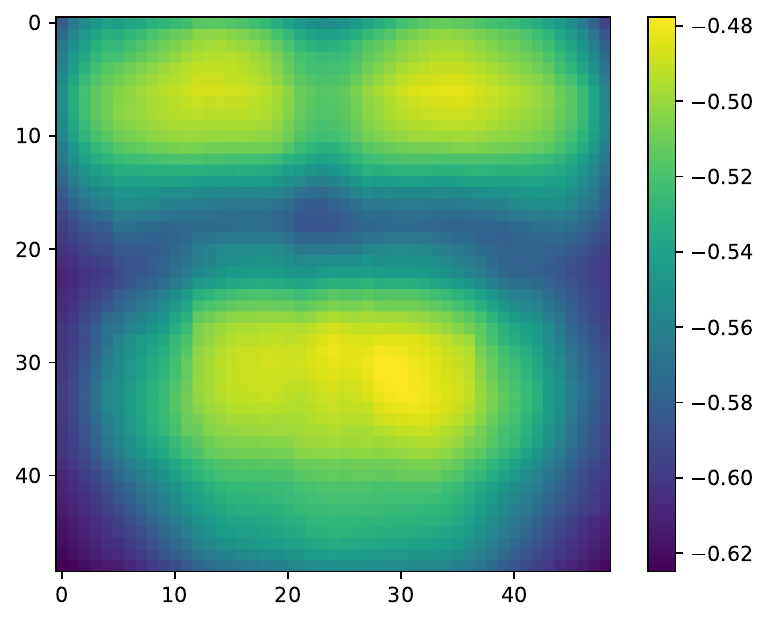}
    \caption{Gender}
    \label{fig:gender}
    \end{subfigure} 
    \begin{subfigure}{0.45\linewidth}
          \includegraphics[width=\linewidth]{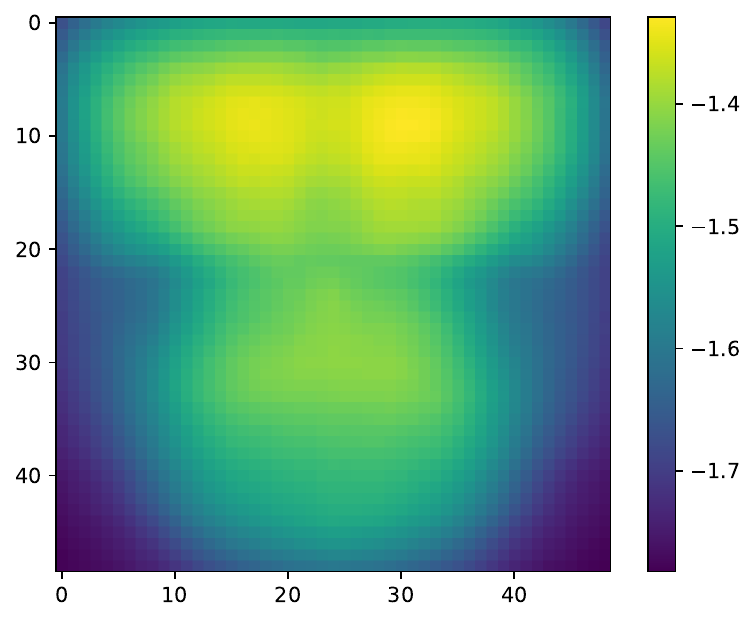}
    \caption{Race}
    \label{fig:race}
    \end{subfigure}
    \begin{subfigure}{0.45\linewidth}
          \includegraphics[width=\linewidth]{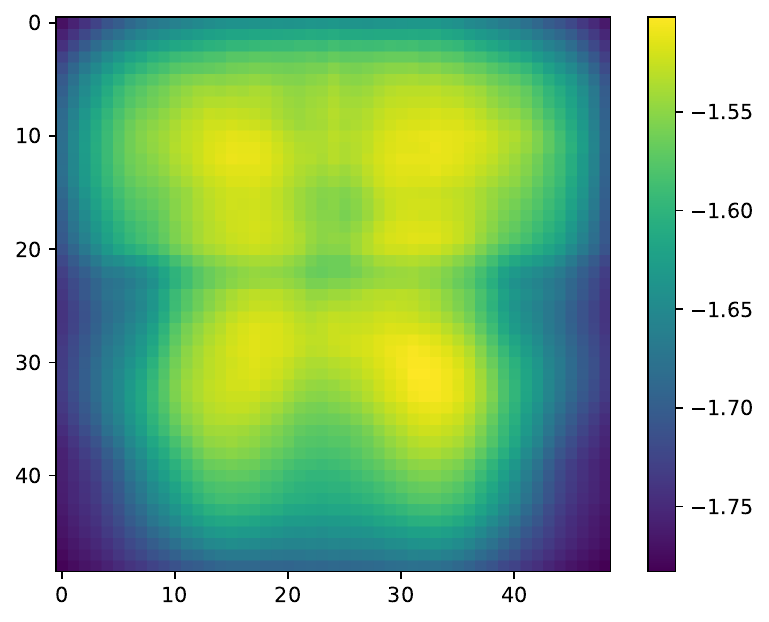}
    \caption{Age}
    \label{fig:age}
    \end{subfigure}
    \caption{Mean attribution maps for the  FairFace attributes.}
    \label{fig:mean_ex}
\end{figure}

\begin{figure}[]
    \centering
    \begin{subfigure}{0.45\linewidth}
          \includegraphics[width=\linewidth]{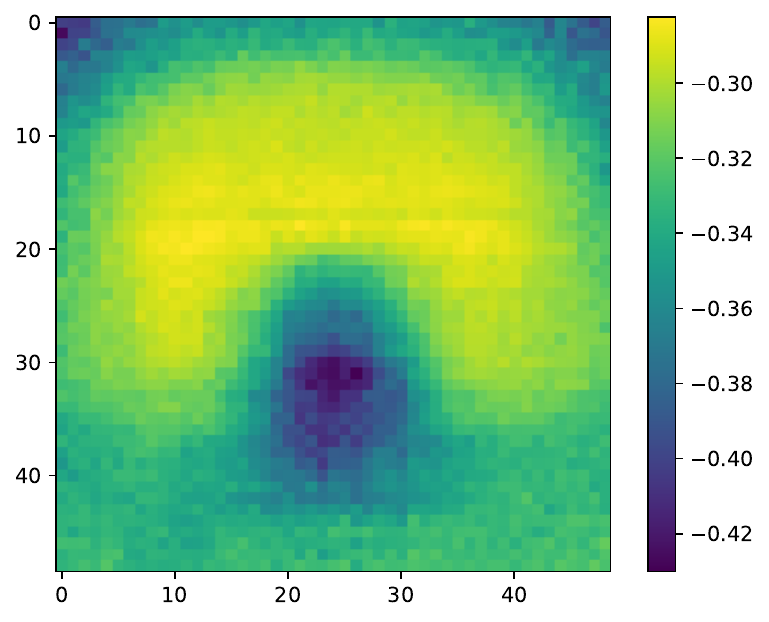}
    \caption{Blond Hair}
    \label{fig:blond_hair}
    \end{subfigure} 
    \begin{subfigure}{0.45\linewidth}
          \includegraphics[width=\linewidth]{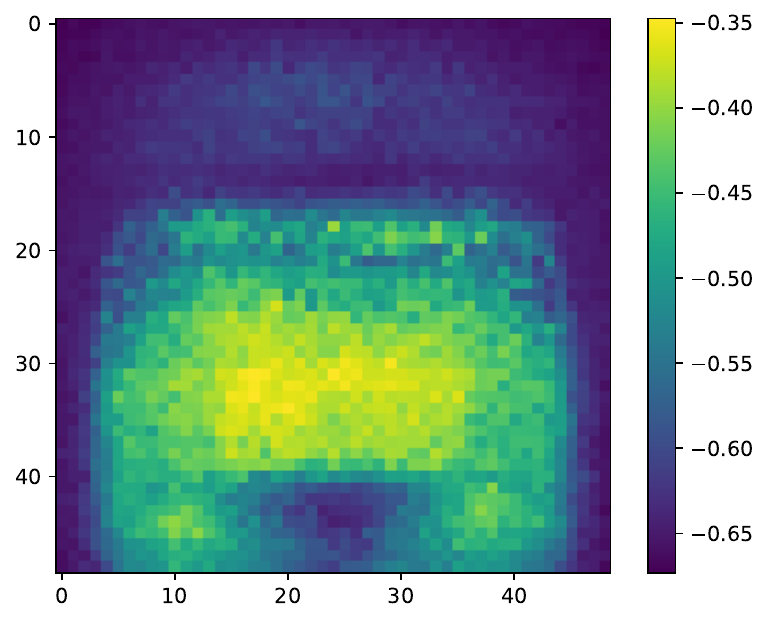}
    \caption{Smiling}
    \label{fig:smiling}
    \end{subfigure}
    \begin{subfigure}{0.45\linewidth}
          \includegraphics[width=\linewidth]{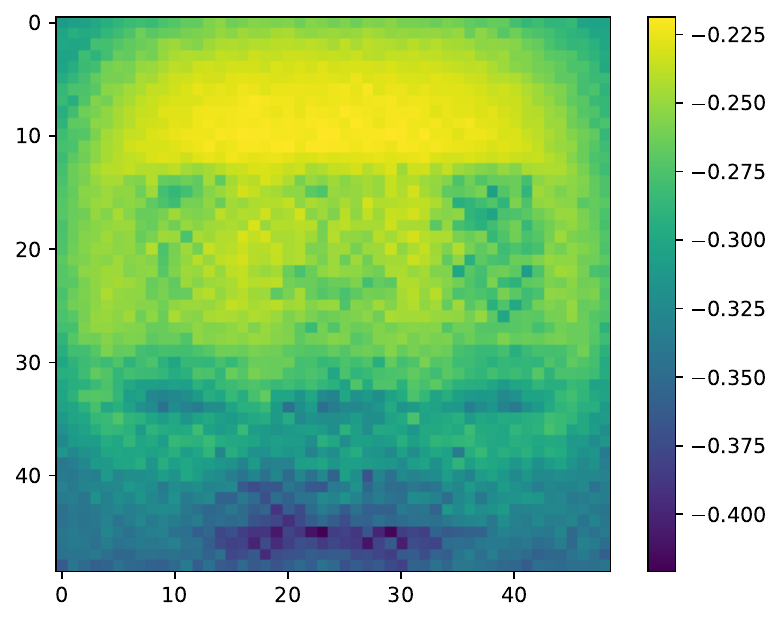}
    \caption{Wearing Hat}
    \label{fig:wearing_hat}
    \end{subfigure}
    \begin{subfigure}{0.45\linewidth}
          \includegraphics[width=\linewidth]{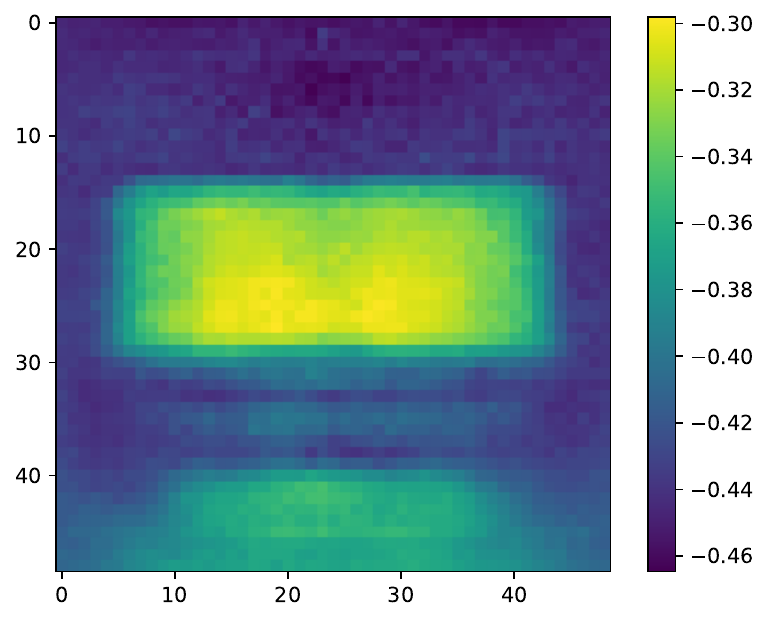}
    \caption{Eyeglasses}
    \label{fig:eyeglasses}
    \end{subfigure}
    \caption{Mean attribution maps for CelebA attributes.}
    \label{fig:mean_all_celeba}
\end{figure}

By averaging over all samples, mean confidence maps can be generated. These maps offer valuable insights into the spatial distribution of important attribute information within an image and can be used to relate these attributes to one another. The calculated maps for the FairFace attributes are presented in Figure \ref{fig:mean_ex}, revealing distinct differences between the attributes. In addition to this, we trained region classifiers on select attributes of the CelebA dataset, for which we can see the mean attributions in Figure \ref{fig:mean_all_celeba}. We observe that the attribution maps for attributes such as Blond Hair and Eyeglasses correspond with human intuition.

For many applications, a pixel-space attribution map is more practical. We can achieve this by converting the region-based map to pixel space using Algorithm \ref{alg:map_attributions_to_pixels}. The algorithm employs the confidences and locations of all patches present on the original image to produce a pixel-level confidence map matching the original image's dimensions. It calculates the confidence for each pixel by averaging the confidences of all patches containing that pixel. For all experiments in Section \ref{sec:fair_training}, we used pixel space attributions, which were normalized to a $[0,1]$ range.

\begin{algorithm}[h]
\caption{Map region attributions to pixel space}
\label{alg:map_attributions_to_pixels}
\begin{algorithmic}[1]
    \STATE \textbf{Input:} $confidences$, $locations$, $k$, $image\_size$
    \STATE $map \gets$ zeros tensor of size $image\_size$
    \STATE $counts \gets$ zeros tensor of size $image\_size$

    \FOR{$i = 0$ \TO length($locations$) - 1}
        \STATE $(x, y) \gets$ $locations[i]$
        \STATE $map[y:y+k, x:x+k] \mathrel{+}= confidences[i]$
        \STATE $counts[y:y+k, x:x+k] \mathrel{+}= 1$
    \ENDFOR
    \RETURN $map / counts$
\end{algorithmic}
\end{algorithm}


%% file: sec/4_attribution.tex
\section{Regularized training on biased data}
\label{sec:fair_training}

To develop a robust training algorithm, we aim to leverage our attribution maps to address biases in the training dataset by introducing targeted noise to regions critical for detecting confounding attributes. Our attribution technique could enable manipulation of the training data to mitigate the model's reliance on these confounders. An illustrative figure can be seen in Figure \ref{fig:region_classifier}.

\subsection{Experiment Setup}

In order to evaluate the feasibility of this approach, we conducted experiments utilizing the CelebA dataset, which comprises a comprehensive collection of annotated facial attributes. Our investigation focuses on a scenario wherein a classifier is trained for predicting perceived gender classification\footnote{In the CelebA dataset, perceived gender is denoted as \textit{Male}, where a value of $1$ represents individuals belonging to the category of men, while a value of $0$ is assigned to those who do not, which we label as women for simplicity's sake.}. The dataset is not sampled uniformly random from the CelebA dataset. Specifically, the dataset is perfectly balanced for the classification attribute (3000 instances of men, 3000 instances of women), but contains bias in terms of another attribute. Within the subset of men, none exhibit the confounding attribute, whereas among women, 2000 individuals possess the confounder while 1000 do not. We created datasets in this manner using Blond Hair, Eyeglasses, Smiling and Wearing Hat as confounding attributes. Subsequently, we assess performance on a test dataset that is balanced with respect to gender as well as the hidden attribute. By introducing this bias into the dataset, our hypothesis posits a significant discrepancy in test accuracy between male and female subjects, owing to the disparate distribution of data instances across gender categories in the test set.

\begin{figure}
    \centering
    \includegraphics[width=\columnwidth]{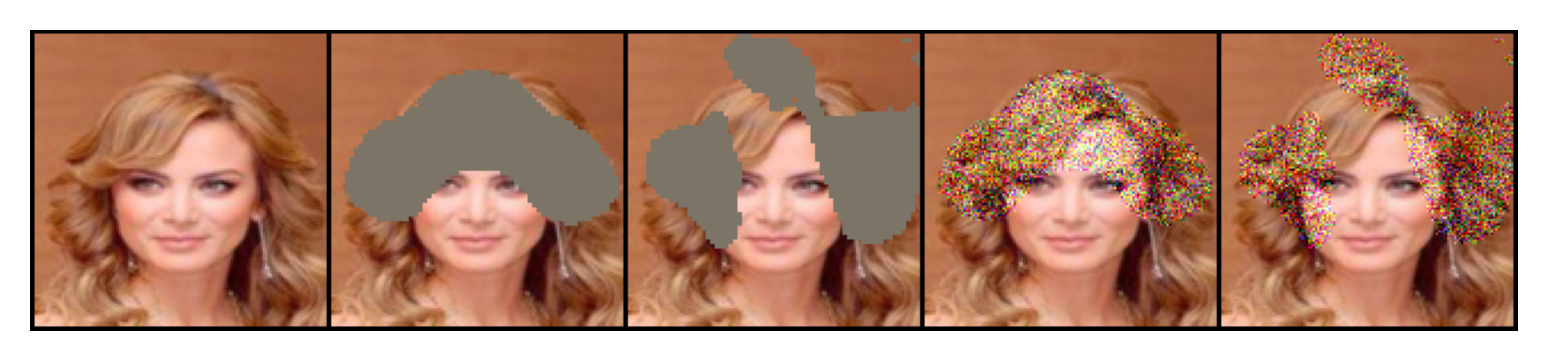}
    \caption{Example of regularizing the data based on attributions for hair color. From left to right: Original, General Mask, Specific Mask, General Noise, Specific Noise.}
    \label{fig:reg_hc}
\end{figure}

To cultivate a robust classifier, we conducted experiments exploring various methods of utilizing attribution maps to alter the training data. We investigated the incorporation of both image-specific and general attributions (calculated over a large number of samples) of confounding attributes, as exemplified in Figure \ref{fig:mean_all_celeba}. For both mask types, we experimented with greying out as well as additive noise. An illustration of adding noise for the Hair Color attribute in all manners (General Mask, Specific Mask, General Noise, Specific Noise) is provided in Figure \ref{fig:reg_hc}. When introducing noise, we drew samples from a Gaussian distribution with a mean of $0$ and a standard deviation of $0.5$. In all cases, we masked out regions deemed most crucial for classifying unintended biases, such as the top $30\%$ most confident regions within an image. The selection of the noise addition scheme and the quantile used for masking out regions were treated as hyperparameters in our analysis.

Several metrics are available for evaluating model fairness, with Demographic Parity and Equality of Opportunity being the most prominent. In our assessment, focusing on perceived gender classification, we deemed it appropriate to examine accuracy per subgroup. This approach allows us to determine if any particular category (in this case, men or women) is disproportionately affected by the known bias.

We employed a customized Convolutional Neural Network (CNN) architecture for gender classification, as we observed that deeper network architectures such as VGG \cite{simonyan2014very} and ResNet \cite{he2016deep} were susceptible to overfitting. Our CNN architecture comprised three convolutional layers, supplemented with dropout regularization \cite{srivastava2014dropout} and batch normalization \cite{ioffe2015batch}. To optimize the model parameters, we utilized the AdamW optimizer \cite{loshchilov2017decoupled} with an initial learning rate set to $10^{-5}$, followed by exponential decay with a decay rate ($\gamma$) of $0.95$. We conducted training with a batch size of $128$. Robust models incorporating noise were trained for $20$ epochs, while those utilizing masking techniques were trained until convergence for $10$ epochs.

\subsection{Results}

\begin{table*}[h]
    \centering
    \resizebox{\linewidth}{!}{
    \begin{tabular}{lllrrrrrrrrrrrr}
\toprule
            &                & {} & \multicolumn{3}{c}{Accuracy $\uparrow$} & \multicolumn{3}{c}{Accuracy Men $\uparrow$} & \multicolumn{3}{c}{Accuracy Women $\uparrow$} & \multicolumn{3}{c}{Gap $\downarrow$} \\
            &                & Type & Original  & Balanced $\Delta$& Ours $\Delta$ &     Original & Balanced $\Delta$& Ours $\Delta$ &       Original & Balanced $\Delta$& Ours $\Delta$& Original & Balanced & Ours \\
Attribute & Noise & Quantile &          &          &        &              &          &        &                &          &        &          &          &        \\
\midrule
Blond Hair & General Mask & 0.60 &     0.74 &    +0.09 &    0.0 &          0.6 &     +0.2 &  +0.02 &           0.88 &    -0.03 &  -0.03 &     0.28 &     0.05 &   0.22 \\
            & General Noise & 0.70 &     0.77 &    +0.06 &  -0.02 &         0.64 &    +0.16 &  +0.08 &            0.9 &    -0.05 &  -0.12 &     0.26 &     0.05 &   0.09 \\
            & Specific Mask & 0.60 &     0.74 &    +0.09 &  +0.01 &          0.6 &     +0.2 &  +0.09 &           0.88 &    -0.03 &  -0.08 &     0.28 &     0.05 &   0.11 \\
            & Specific Noise & 0.80 &     0.79 &    +0.04 &  -0.05 &         0.67 &    +0.13 &  +0.01 &            0.9 &    -0.05 &   -0.1 &     0.23 &     0.05 &   0.15 \\
Eyeglasses & General Mask & 0.60 &     0.71 &    +0.06 &  -0.02 &         0.61 &    +0.11 &  +0.03 &           0.82 &     +0.0 &  -0.07 &     0.21 &     0.10 &   0.12 \\
            & General Noise & 0.60 &     0.72 &    +0.05 &  +0.01 &         0.61 &    +0.11 &  +0.05 &           0.84 &    -0.02 &  -0.04 &     0.23 &     0.10 &   0.14 \\
            & Specific Mask & 0.60 &     0.71 &    +0.06 &  +0.03 &         0.61 &    +0.11 &  +0.13 &           0.82 &     +0.0 &  -0.08 &     0.21 &     0.10 &   0.05 \\
            & Specific Noise & 0.80 &     0.72 &    +0.05 &  -0.01 &         0.61 &    +0.11 &   +0.1 &           0.84 &    -0.02 &  -0.13 &     0.23 &     0.10 &   0.09 \\
Smiling & General Mask & 0.80 &     0.81 &    +0.03 &  -0.04 &         0.71 &    +0.06 &  -0.02 &            0.9 &    +0.01 &  -0.05 &     0.19 &     0.13 &   0.16 \\
            & General Noise & 0.60 &     0.85 &    -0.01 &  -0.05 &         0.79 &    -0.02 &  -0.04 &           0.91 &     -0.0 &  -0.07 &     0.12 &     0.13 &   0.09 \\
            & Specific Mask & 0.95 &     0.81 &    +0.03 &  -0.02 &         0.71 &    +0.06 &  -0.02 &            0.9 &    +0.01 &  -0.01 &     0.19 &     0.13 &   0.20 \\
            & Specific Noise & 0.60 &     0.85 &    -0.01 &  -0.05 &         0.79 &    -0.02 &  +0.01 &           0.91 &     -0.0 &   -0.1 &     0.12 &     0.13 &   0.05 \\
Wearing Hat & General Mask & 0.80 &     0.71 &    +0.08 &  +0.05 &         0.63 &    +0.11 &  +0.08 &           0.79 &    +0.05 &  +0.03 &     0.16 &     0.10 &   0.10 \\
            & General Noise & 0.95 &     0.73 &    +0.06 &  -0.02 &         0.63 &    +0.11 &  +0.03 &           0.84 &     +0.0 &  -0.07 &     0.21 &     0.10 &   0.11 \\
            & Specific Mask & 0.95 &     0.71 &    +0.08 &    0.0 &         0.63 &    +0.11 &  +0.01 &           0.79 &    +0.05 &  -0.01 &     0.16 &     0.10 &   0.15 \\
            & Specific Noise & 0.95 &     0.73 &    +0.06 &  -0.03 &         0.63 &    +0.11 &  +0.04 &           0.84 &     +0.0 &   -0.1 &     0.21 &     0.10 &   0.08 \\
\bottomrule
\end{tabular}
}
    \caption{Summary of the noise addition experiment across multiple attributes of the CelebA dataset. Models are trained on a highly biased dataset regarding each attribute, leading to a disparity in performance between men and women. This disparity between performance for men and women is denoted in the gap column. Accuracies for models trained on a balanced dataset, and those using our noise addition regularization are shown relative to those trained on the original, biased dataset. Through the strategic addition of noise to regions crucial for each attribute based on a specified noise type, we diminish the discrepancy in accuracies between genders.}
    \label{tab:top_results}
\end{table*}

We conducted experiments using all attributes illustrated in Figure \ref{fig:mean_all_celeba} as confounding attributes, with the results summarized in Table \ref{tab:top_results}. The table presents overall accuracy, as well as accuracy for men and women separately on a balanced test set, with the gap column indicating the difference between these two. Notably, our experiments encompassed multiple quantiles, yet we only included those yielding the most significant results for brevity. Additionally, we compared these results against classifiers trained on a dataset of equivalent size but balanced with respect to the attribute. It's worth noting that the data was not fully balanced in the case of Blond Hair as the confounder for men but rather $54/46$ in favor of non-blond hair, as the original CelebA train set contains only $1387$ men with blond hair.

First of all, we can observe that all models have some discrepancy between the accuracy for men and women, with those trained on the biased dataset without any modifications having the largest gap. Nevertheless, we observe that for each confounding attribute, there exists a specific combination of noise type and quantile for regularizing that yields notably improved equality in accuracy, despite the significant divergence in the training distribution of men compared to the balanced test data. Surprisingly, even when trained on balanced data concerning the attribute, we still observe a performance gap between men and women, albeit significantly smaller than with highly unbalanced data. This discrepancy likely persists due to other confounders that remain present, as we did not balance the data on the remaining $38$ attributes. Furthermore, in the case of smiling, there is almost no discernible difference between balanced and unbalanced data. This observation may be attributed to the inherent difficulty in classifying smiling compared to gender, resulting in a lesser impact of this bias on the model~\cite{hall2022systematic}. However, our technique notably reduces the accuracy gap between men and women, approaching the performance achieved with balanced data, suggesting that our method is a more practical approach than obtaining a balanced dataset on all attributes.

\begin{figure}[h]
    \centering
    \includegraphics[width=\columnwidth]{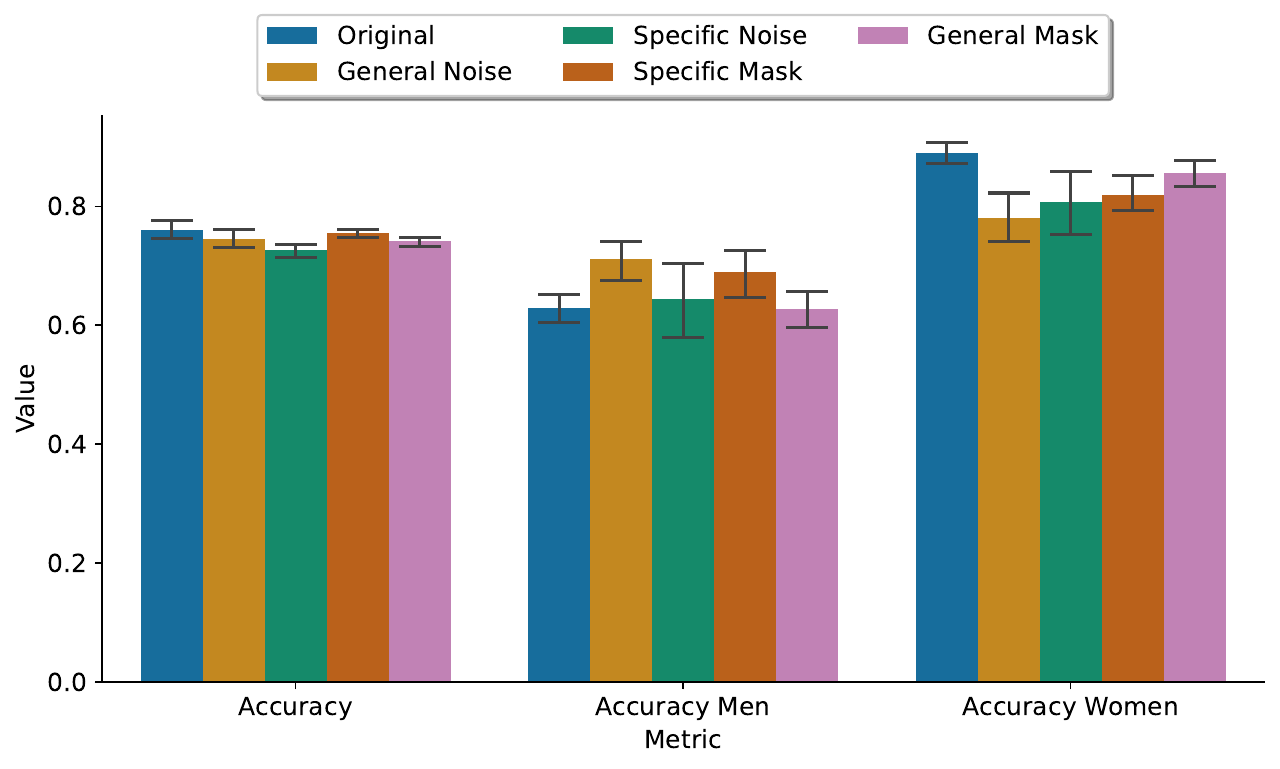}
    \caption{Comparison of noise addition techniques for the attribute Blond Hair using Quantile $0.7$}
    \label{fig:blond_q70}
\end{figure}

Subsequently, we zoom in on the choice of the two hyperparameters. Figure \ref{fig:blond_q70} illustrates a comparison of various noise addition strategies for the Blond Hair attribute, where we removed the $30\%$ most influential information related to the attribute from all training data. All robust training schemes either enhance or maintain the accuracy for men, while diminishing the accuracy for women, thus aligning subgroup accuracies and mitigating the effect of training on a biased dataset. Notably, the incorporation of General Noise brings both subset accuracies closest to parity. 

\begin{figure}[h]
    \centering
    \includegraphics[width=\columnwidth]{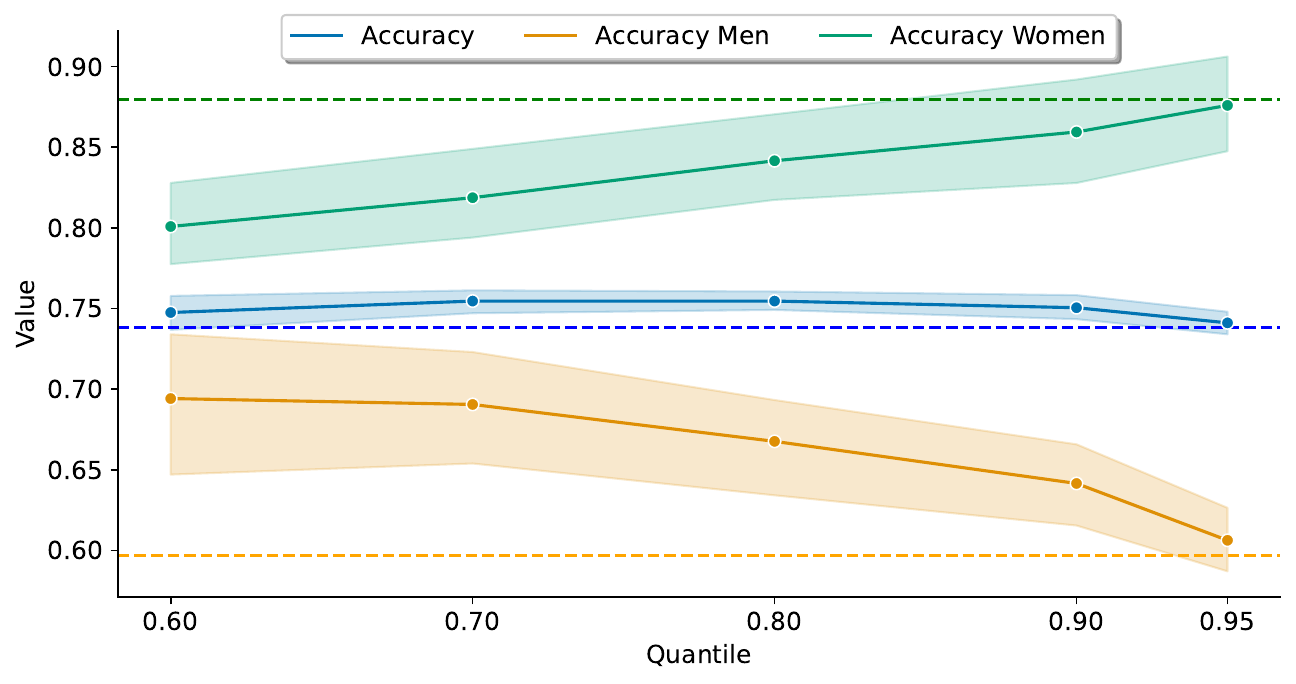}
    \caption{Comparison of metrics for the Specific Mask noise addition, for the attribute Blond Hair. Dotted lines indicate the accuracy of the original model for the same metric.}
    \label{fig:quantile_comparison}
\end{figure}

Next to the type of noise, the quantile of information blocked is an important hyperparameter as well. Figure \ref{fig:quantile_comparison} shows how the accuracy of the subgroups evolves with the differing percentages. In the case of Specific Mask for the Blond Hair attribute, the optimal point seems to be the $70\%$ mark. Masking out more reduces accuracy for women further while more specific noise additions increase the inequality in performance to that of unmodified data. 

%% file: sec/5_conclusion.tex
\section{Conclusion}
\label{sec:conclusion}
In conclusion, our paper introduces a novel method for mitigating bias in machine learning models, crucial for ensuring fairness and equity. We propose leveraging pixel image attributions to identify and regulate regions within images containing significant information about biased attributes. Our approach, employing a model-agnostic technique to extract pixel attributions through a CNN classifier trained on small image patches, enables the identification of critical information distribution across images. By utilizing these attributions to introduce targeted noise into datasets with confounding attributes, we effectively prevent neural networks from learning biases and prioritize primary attributes. Our method demonstrates its effectiveness in training unbiased classifiers on heavily biased datasets, offering promise for enhancing fairness and equity in machine learning applications.

%% file: sec/ack.tex
\section*{Acknowledgments}

Sander De Coninck receives funding from the Special Research Fund of Ghent University under grant no. BOF22/DOC/093. This research received funding from the Flemish Government under the “Onderzoeksprogramma Artificiële Intelligentie (AI) Vlaanderen” programme.